\pdfoutput=1
\documentclass[runningheads]{llncs}
\usepackage{graphicx}
\usepackage{amsmath,amssymb} 
\usepackage{color}
\graphicspath{ {images/} }
\usepackage{multirow}
\usepackage{multicol}
\usepackage[colorlinks,
linkcolor=red,
anchorcolor=green,
citecolor=blue
]{hyperref}

\usepackage{bbding}
\begin{document}
\title{Hierarchical Bilinear Pooling \\ for Fine-Grained Visual Recognition}

\titlerunning{Hierarchical Bilinear Pooling for Fine-Grained Visual Recognition}
%
\author{Chaojian Yu \and
Xinyi Zhao \and
Qi Zheng \and
Peng Zhang \and
Xinge You$^{(}$\Envelope$^{)}$
}
%
\authorrunning{C. Yu et al.}
%

\institute{School of Electronic Information and Communications\\
Huazhong University of Science and Technology, Wuhan, China\\
\email{\{yucj,youxg\}@hust.edu.cn}}
\maketitle              
\begin{abstract}
Fine-grained visual recognition is challenging because it highly relies on the modeling of various semantic parts and fine-grained feature learning. Bilinear pooling based models have been shown to be effective at fine-grained recognition, while most previous approaches neglect the fact that inter-layer part feature interaction and fine-grained feature learning are mutually correlated and can reinforce each other. In this paper, we present a novel model to address these issues. First, a cross-layer bilinear pooling approach is proposed to capture the inter-layer part feature relations, which results in superior performance compared with other bilinear pooling based approaches. Second, we propose a novel hierarchical bilinear pooling framework to integrate multiple cross-layer bilinear features to enhance their representation capability. Our formulation is intuitive, efficient and achieves state-of-the-art results on the widely used fine-grained recognition datasets.

\end{abstract}
\section{Introduction}
With the development of artificial intelligence, increasing demand appears to recognize subcategories of objects under the same basic-level category, e.g., brand identification for businessman, plant recognition for botanist. Thus recent years have witnessed great progress in fine-grained visual recognition, which has been widely used in applications such as automatic driving~\cite{sochor2016boxcars}, expert-level image recognition~\cite{krause2016unreasonable}, etc. Different from general image classification task (e.g., ImageNet classification~\cite{russakovsky2015imagenet}) that is to distinguish basic-level categories, fine-grained visual recognition is very challenging as subcategories tend to own small variance in object appearance and thus can only be recognized by some subtle or local differences. For example, we discriminate breeds of birds depending on the color of their back or the shape of their beak.

Motivated by the observation that local parts of object usually act a role of importance in differentiating subcategories, many methods~\cite{zhang2016spda,branson2014bird,simon2015neural,zhang2016picking} for fine-grained classification were developed by exploiting the parts, namely part-based approaches. They mainly consist of two steps: firstly localize the foreground object or object parts, e.g., by utilizing available bounding boxes or part annotations, and then extract discriminative features for further classification. However, these approaches suffer from two essential limitations. First, it is difficult to ensure the manually defined parts are optimal or suitable for the final fine-grained classification task. Second, detailed part annotations incline to be time consuming and labor intensive, which is not feasible in practice. Therefore, some other approaches employ unsupervised techniques to detect possible object regions. For example, Simon and Rodner~\cite{simon2015neural} proposed a constellation model to localize parts of objects, leveraging convolutional neural network (CNN) to find the constellations of neural activation patterns. Zhang \emph{et al.}~\cite{zhang2016picking} proposed an automatic fine-grained image classification method, incorporating deep convolutional filters for both selection and description related to parts. These models regard CNN as part detector and obtain great improvement in fine-grained recognition. Unlike part-based methods, we treat activations from different convolution layers as responses to different part properties instead of localizing object parts explicitly, leveraging cross-layer bilinear pooling to capture inter-layer interaction of part attributes, which is proved to be useful for fine-grained recognition.

Alternatively, some researches~\cite{cai2017higher,gao2016compact,lin2015bilinear,kong2017low} introduced bilinear pooling frameworks to model local parts of object. Although promising results have been reported, further improvement suffers from the following limitations. First, most existing bilinear pooling based models only take activations of the last convolution layer as representation of an image, which is insufficient to describe various semantic parts of object. Second, they neglect intermediate convolution activations, resulting in a loss of discriminative information of fine-grained categories which is significant for fine-grained visual recognition.

In this work, we present new methods to address the above challenges. We find that inter-layer part feature interaction and fine-grained feature learning are mutually correlated and can reinforce each other. To better capture the inter-layer feature relations, we propose a cross-layer bilinear pooling approach. The proposed method is efficient and powerful. It takes into account the inter-layer feature interactions while avoiding introducing extra training parameters. In contrast to other bilinear pooling based works which only utilize feature from one single convolution layer, our architecture exploits the interaction of part features from multiple layers, which is useful for fine-grained feature learning. Besides, our framework is highly consistent with the human coarse-to-fine perception, the visual hierarchy segregates local and global features in cortical areas V4 based on spatial differences and builds a temporal dissociation of the neural activity~\cite{lu2018revealing}. We find that our cross-layer bilinear model is closer to the unique architecture of cortical areas V4 for processing spatial information.

It is well known that information loss exists in the propagation of CNNs. In order to minimize the loss of information that is useful for fine-grained recognition, we propose a novel hierarchical bilinear pooling framework to integrate multiple cross-layer bilinear features to enhance their representation power. To make full use of the intermediate convolution layer activations, all cross-layer bilinear features are concatenated before the final classification. Note that the features from different convolution layer are complementary, they contribute to discriminative feature learning. Thus the proposed network benefits from the mutual reinforcement between inter-layer feature interaction and fine-grained feature learning. Our contributions are summarized as follows:

\begin{itemize}
\item[\textbullet] We develop a simple but effective cross-layer bilinear pooling technique that simultaneously enables the inter-layer interaction of features and the learning of fine-grained representation in a mutually reinforced way.
\item[\textbullet] Based on cross-layer bilinear pooling, we propose a hierarchical bilinear pooling framework to integrate multiple cross-layer bilinear modules to obtain the complementary information from intermediate convolution layers for performance boost.
\item[\textbullet] We conduct comprehensive experiments on three challenging datasets (CUB Birds, Stanford Cars, FGVC-Aircraft), and the results demonstrate the superiority of our method.
\end{itemize}

The rest of this paper is organized as follows. Section~\ref{sec:re_w} reviews the related work. Section~\ref{sec:model} presents the proposed method. Section~\ref{sec:exp} provides experiments as well as result analysis, followed by conclusion in Section~\ref{sec:conc}.

\section{Related Work}
\label{sec:re_w}
In the following, we briefly review previous works from the two viewpoints of interest due to their relevance to our work, including fine-grained feature learning and feature fusion in CNNs.

\subsection{Fine-Grained Feature Learning}
Feature learning plays an important and fundamental role in fine-grained recognition. Since the differences between subcategories are subtle and local, capturing global semantic information with merely fully connected layers limits the representation capacity of a framework, and hence restricts further promotion of final recognition~\cite{babenko2015aggregating}. To better model subtle difference for fine-grained categories, Lin \emph{et al.}~\cite{lin2015bilinear} proposed a bilinear structure to aggregate the pairwise feature interactions by two independent CNNs, which adopted outer product of feature vectors to produce a very high-dimensional feature for quadratic expansion. Gao \emph{et al.}~\cite{gao2016compact} applied Tensor Sketch~\cite{pham2013fast} to approximate the second-order statistics and to reduce feature dimension. Kong \emph{et al.}~\cite{kong2017low} adopted low-rank approximation to the covariance matrix and further reduced the computational complexity. Yin \emph{et al.}~\cite{cui2017kernel} aggregated higher-order statistics by iteratively applying the Tensor Sketch compression to the features. The work in~\cite{moghimi2016boosted} utilized bilinear convolutional neural network as baseline model and adopted an ensemble learning method to incorporate boosting weights. In~\cite{lin2017improved}, matrix square-root normalization was proposed and proved to be complementary to existing normalization. However, these approaches only consider the feature from single convolution layer, which is insufficient to capture various discriminative parts of object and model the subtle differences among subcategories. The method we propose overcome this limitation via integrating inter-layer feature interaction and fine-grained feature learning in a mutually reinforced manner and is therefore more effective.

\subsection{Feature Fusion in CNNs}
Due to the success of deep learning, CNNs have emerged as general-purpose feature extractors for a wide range of visual recognition tasks. While feature maps from single convolution layer are insufficient for finer-grained tasks, thus some recent works~\cite{cai2017higher,hariharan2015hypercolumns,long2015fully,xie2015holistically} attempt to investigate the effectiveness of exploiting feature from different convolution layers within a CNN. For example, Hariharan \emph{et al.}~\cite{hariharan2015hypercolumns} considered the feature maps from all convolution layers, allowing finer grained resolution for localization tasks. Long \emph{et al.}~\cite{long2015fully} combined the finer-level and higher-level semantic feature from different convolution layers for better segmentation. Xie \emph{et al.}~\cite{xie2015holistically} proposed a holistically-nested framework where the side outputs are added after lower convolution layers to provide deep supervision for edge detection. The very recent work~\cite{cai2017higher} concatenated the activation maps from multiple convolution layers to model the interaction of part features for fine-grained recognition. However, simply cascading the feature map introduces lots of training parameters and even fails to capture inter-layer feature relations when incorporating with more intermediate convolution layers. Instead, our network treats each convolution layer as attribute extractor for different object parts and models their interactions in an intuitive and effective way.

\section{Hierarchical Bilinear Model}
\label{sec:model}
In this section, we develop a hierarchical bilinear model to overcome those limitations mentioned above. Before presenting our hierarchical bilinear model, we first introduce the general formulation of factorized bilinear pooling for fine-grained image recognition in Sect.~\ref{sec:model_fbp}. Based on this, we propose a cross-layer bilinear pooling technique to jointly learn the activations from different convolution layers in Sect.~\ref{sec:model_cbp}, which captures the cross-layer interaction of information and leads to better representation capability. Finally, our hierarchical bilinear model combining multiple cross-layer bilinear modules generates finer part description for better fine-grained recognition in Sect.~\ref{sec:model_hbm}.

\subsection{Factorized Bilinear Pooling} \label{sec:model_fbp}
Factorized bilinear pooling has been applied to visual question answer task, Kim \emph{et al.}~\cite{kim2016hadamard} proposed factorized bilinear pooling using Hadamard product for an efficient attention mechanism of multimodal learning. Here we introduce the basic formulation of factorized bilinear pooling technique for the task of fine-grained image recognition. Suppose an image $I$ is filtered by a CNN and the output feature map of a convolution layer is $X \in \mathbb{R}^{h \times w \times c}$ with height $h$, width $w$ and channels $c$, we denote a $c$ dimensional descriptor at a spatial location on $X$ as $\mathbf{x} = [x_1,x_2,\cdots,x_c]^T$. Then the full bilinear model is defined by
\begin{align}
\label{eq:eq1}
  z_i=\mathbf{x}^TW_i\mathbf{x}
\end{align}
Where $W_i \in \mathbb{R}^{c \times c}$ is a projection matrix, $z_i$ is the output of the bilinear model. We need to learn $\mathbf{W}=[W_1,W_2,\cdots,W_o] \in \mathbb{R}^{c \times c \times o}$ to obtain a $o$ dimensional output $\mathbf{z}$. According to matrix factorization in~\cite{rendle2010factorization}, the projection matrix $W_i$ in Eq.~(\ref{eq:eq1}) can be factorized into two one-rank vectors
\begin{align}
\label{eq:eq2}
  z_i=\mathbf{x}^TW_i\mathbf{x}=\mathbf{x}^TU_iV_i^T\mathbf{x}=U_i^T\mathbf{x} \circ V_i^T\mathbf{x}
\end{align}
where $U_i \in \mathbb{R}^c$ and $V_i \in \mathbb{R}^c$. Thus the output feature $\mathbf{z} \in \mathbb{R}^o$ is given by
\begin{align}
\label{eq:eq3}
  \mathbf{z}=P^T(U^T\mathbf{x} \circ V^T\mathbf{x})
\end{align}
where $U \in \mathbb{R}^{c \times d}$ and $V \in \mathbb{R}^{c \times d}$ are projection matrices, $P \in \mathbb{R}^{d \times o}$ is the classification matrix, $\circ$ is the Hadamard product and $d$ is a hyperparameter deciding the dimension of joint embeddings.

\subsection{Cross-Layer Bilinear Pooling} \label{sec:model_cbp}
Fine-grained subcategories tend to share similar appearances and can only be discriminated by subtle differences in the attributes of local part, such as color, shape, or length of beak for birds. Bilinear pooling, which captures the pairwise feature relations, is an important technique for fine-grained recognition. However, most bilinear models only focus on learning the features from single convolution layer while completely ignoring the cross-layer interaction of information. Activations of individual convolution layer are incomplete since there are multiple attributes in each object part which can be crucial in differentiating subcategories.

Actually in most cases, we need to simultaneously consider multi-factor of part feature to determine the category for a given image. Therefore, to capture finer grained part feature, we develop a cross-layer bilinear pooling approach that treats each convolution layer in a CNN as part attributes extractor. After that the features from different convolution layers are integrated by element-wise multiplication to model the inter-layer interaction of part attributes. Accordingly, Eq.~(\ref{eq:eq3}) can be rewritten as
\begin{align}
  \mathbf{z}=P^T(U^T\mathbf{x} \circ V^T\mathbf{y})
\end{align}
where $\mathbf{x}$ and $\mathbf{y}$ represent local descriptors from different convolution layers at the same spatial location.

It is worth noting that the features from different convolution layers are expanded into high-dimensional space by independent linear mappings. It is expected that the convolution activations and project activations encode global and local feature of object respectively, as shown in Fig.~\ref{fig:fig3}. It is highly consistent with the human coarse-to-fine perception: human and non-human primates often see the global ``gist'' of an object, or a scene, before discerning local detailed features~\cite{lu2018revealing}. For example, neurons in macaque inferotemporal cortex that are active during face perception encode the global facial category is earlier than they begin to encode finer information such as identity or expression.

\begin{figure}
\centering
\includegraphics[height=6.5cm]{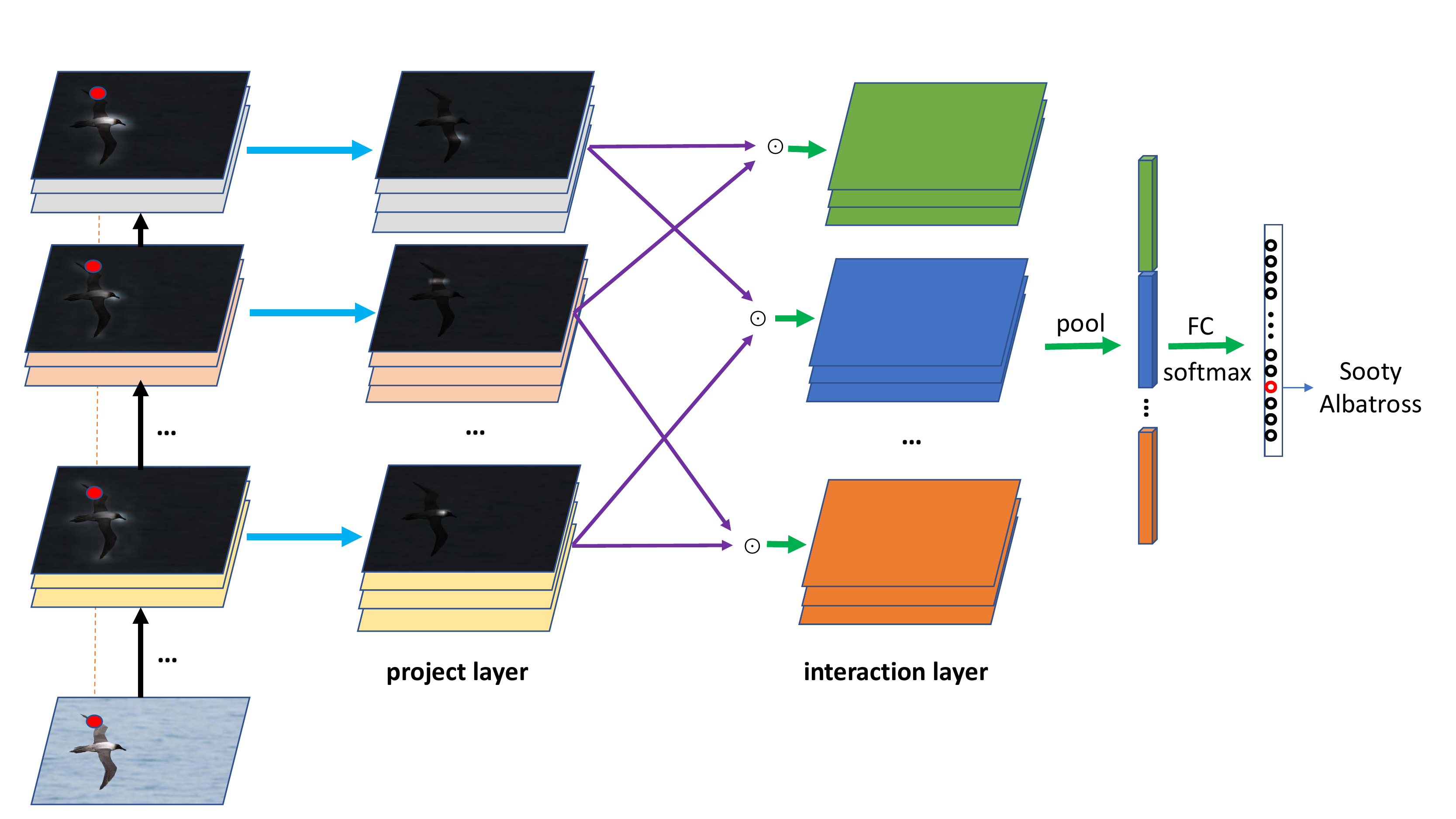}
\caption{Illustration of our Hierarchical Bilinear Pooling (HBP) network architecture for fine-grained recognition. The bottom image is the input, and above it are the feature maps of different layers in the CNN. First the features from different layers are expanded into a high-dimensional space via independent linear mapping to capture attributes of different object parts and then integrated by element-wise multiplication to model the inter-layer interaction of part attributes. After that sum pooling is performed to squeeze the high-dimensional features into compact ones. Note that we obtain the visual activation maps above by computing the response of sum-pooled feature vector on every single spatial location.
}
\label{fig:fig1}
\end{figure}

\subsection{Hierarchical Bilinear Pooling} \label{sec:model_hbm}
Cross-layer bilinear pooling proposed in Sect.~\ref{sec:model_cbp} is intuitive and effective, as it has superior representation capacity than traditional bilinear pooling models without increasing training parameters. This inspires us that exploiting the inter-layer feature interactions among different convolution layers is beneficial for capturing the discriminative part properties between fine-grained subcategories. Therefore, we extend the cross-layer bilinear pooling to integrate more intermediate convolution layers, which further enhances the representation capacity of features. In this section, we propose a generalized Hierarchical Bilinear Pooling (HBP) framework to incorporate more convolutional layer features by cascading multiple cross-layer bilinear pooling modules.

Specifically, we divide the cross-layer bilinear pooling module into interaction stage and classification stage, which formulates as follows
\begin{align}
  \mathbf{z}_{int}=U^T\mathbf{x} \circ V^T\mathbf{y}\\
  \mathbf{z}=P^T\mathbf{z}_{int} \in \mathbb{R}^o
\end{align}
To better model inter-layer feature interactions, the interaction feature of the HBP model is obtained by concatenating multiple $\mathbf{z}_{int}$ of the cross-layer bilinear pooling modules. Thus we can derive final output of the HBP model by
\begin{align}
  \mathbf{z}_{HBP}&=HBP(\mathbf{x},~\mathbf{y},~\mathbf{z},~\cdots)=P^T\mathbf{z}_{int}\\
  &=P^Tconcat(U^T\mathbf{x} \circ V^T\mathbf{y},~U^T\mathbf{x} \circ S^T\mathbf{z},~V^T\mathbf{y} \circ S^T\mathbf{z},~\cdots)
\end{align}
where $P$ is the classification matrix, $U,V,S,\ldots$ are the projection matrices of convolution layer feature $\mathbf{x},\mathbf{y},\mathbf{z},\ldots$ respectively. The overall flowchart of the HBP framework is illustrated in Fig.~\ref{fig:fig1}.

\section{Experiments}
\label{sec:exp}
In this section, we evaluate the performance of HBP model for fine-grained recognition. The datasets and implementation details of HBP are firstly introduced in Sect.~\ref{sec:dataset}. Model configuration studies are performed to investigate the effectiveness of each component in Sect.~\ref{sec:config}. Comparison with state-of-the-art methods is provided in Sect.~\ref{sec:comp}. Finally in Sect.~\ref{sec:qual}, qualitative visualization is present to intuitively explain our model.

\subsection{Datasets and Implementation Details} \label{sec:dataset}
\subsubsection{Datasets:}
We conduct experiments on three widely used datasets for fine-grained image recognition, including Caltech-UCSD Birds (CUB-200-2011)~\cite{wah2011caltech}, Stanford Cars~\cite{krause20133d} and FGVC-Aircraft~\cite{maji2013fine}. The detailed statistics with category numbers and data splits are summarized in Table~\ref{table:table1}. Note that we only use category labels in our experiments.
\begin{table}
\begin{center}
\caption{Summary statistics of datasets}
\label{table:table1}
\begin{tabular}{|c|c|c|c|}
\hline
Datasets & \#Category & \#Training & \#Testing \\
\hline\hline
CUB-200-2011~\cite{wah2011caltech}  & 200 & 5994 & 5794 \\
Stanford Cars~\cite{krause20133d} & 196 & 8144 & 8041 \\
FGVC-Aircraft~\cite{maji2013fine} & 100 & 6667 & 3333 \\
\hline
\end{tabular}
\end{center}
\end{table}
\subsubsection{Implementation Detail:}
For fair comparison with other state-of-the-art methods, we evaluate our HBP with VGG-16~\cite{simonyan2014very} baseline model pretrained on ImageNet classification dataset~\cite{russakovsky2015imagenet}, removing the last three fully-connected layers and inserting all the components in our framework. It is worth noting that our HBP can be also applied to other network structures, such as Inception~\cite{szegedy2015going} and ResNet~\cite{he2016deep}. The size of input image is $448 \times 448$. Our data augmentation follows the commonly used practice, i.e., random sampling (crop $448 \times 448$ from $512 \times S$ where $S$ is the largest image side) and horizontal flipping are utilized during training, and only center cropping is involved during inference.

We initially train only the classifiers by logistic regression, and then fine-tune the whole network using stochastic gradient descent with a batch size of 16, momentum of 0.9, weight decay of $5 \times 10^{-4}$ and a learning rate of $10^{-3}$, periodically annealed by 0.5. All experiments are implemented with the Caffe toolbox~\cite{jia2014caffe} and performed on a server with Titan X GPUs. The source code and trained model will be made available at \url{https://github.com/ChaojianYu/Hierarchical-Bilinear-Pooling}

\subsection{Configurations of Hierarchical Bilinear Pooling} \label{sec:config}
\begin{figure}
\centering
\includegraphics[height=6.5cm]{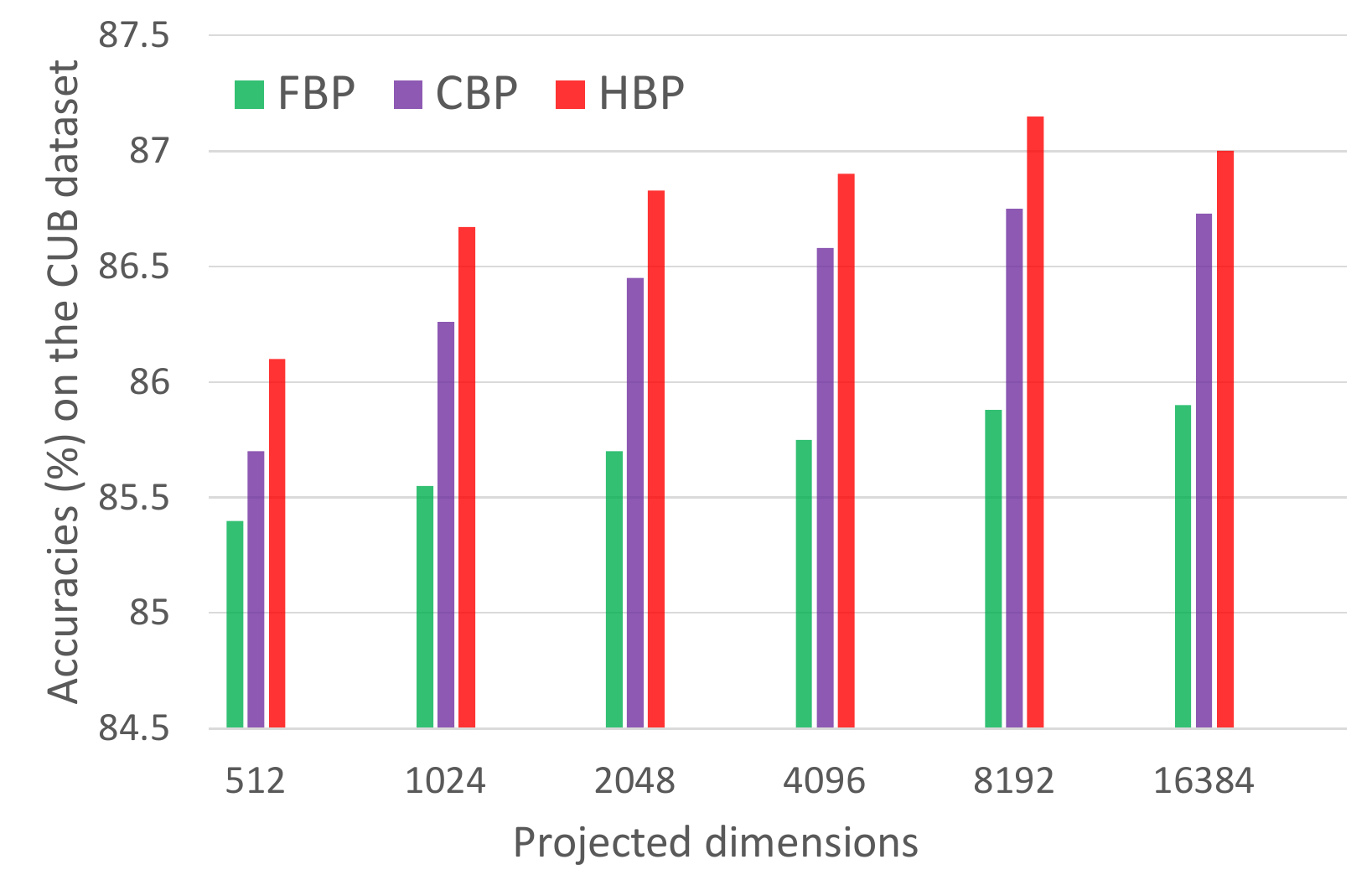}
\caption{Classification accuracy on the CUB dataset. Comparison of general Factorized Bilinear Pooling (FBP), Cross-layer Bilinear Pooling (CBP) and Hierarchical Bilinear Pooling (HBP) with various projection dimensions.}
\label{fig:fig2}
\end{figure}

Cross-layer bilinear pooling (CBP) has a user-define projection dimension $d$. To investigate the impact of $d$ and to validate the effectiveness of the proposed framework, we conduct extensive experiments on the CUB-200-2011~\cite{wah2011caltech} dataset, with results summarized in Fig.~\ref{fig:fig2}. Note that we utilize $relu5\_3$ in FBP, $relu5\_2$ and $relu5\_3$ in CBP, $relu5\_1,~relu5\_2$ and $relu5\_3$ in HBP to obtain the results in Fig.~\ref{fig:fig2} and we also provide quantitative experiments about the choice of layers in the following. We focus on $relu5\_1,~relu5\_2$ and $relu5\_3$ in VGG-16~\cite{simonyan2014very} as they contain more part semantic information compared with shallower layers.

In Fig.~\ref{fig:fig2}, we compare the performance of CBP with the general factorized bilinear pooling model, namely FBP. Futhermore, we explore HBP with combination of multiple layers. Finally, we analyze the impact factors of hyperparameter $d$. We can draw the following significant conclusions from Fig.~\ref{fig:fig2}

\begin{itemize}
\item[\textbullet] First, under the same $d$, our CBP significantly outperforms FBP, which indicates that the discriminative power can be enhanced by the inter-layer interaction of features.
\item[\textbullet] Second, HBP further outperforms CBP, which demonstrates the efficacy of activations from intermediate convolution layers for fine-grained recognition. This can be explained by the fact that information loss exists in the propagation of CNNs, thus discriminative features crucial for fine-grained recognition may be lost in intermediate convolution layers. In contrast to CBP, our HBP takes more feature interactions of intermediate convolution layers into consideration and is therefore more robust, since HBP has presented the best performance. In the following experiments, HBP is used to compare with other state-of-the-art methods.
\item[\textbullet] Third, when $d$ varies from 512 to 8192, increasing $d$ leads to higher accuracy for all models and HBP is saturated with $d=8192$. Therefore, $d=8192$ is used for HBP in our following experiments in consideration of feature dimension, computational complexity as well as accuracy.
\end{itemize}

We then provide quantitative experiments on the CUB-200-2011~\cite{wah2011caltech} dataset to analyze the impact factor of layers. The accuracies in Table~\ref{table:table11} are obtained under the same embedding dimension ($d=8192$). We consider the combination of different layers for CBP and HBP. The results demonstrate that the performance gain of our framework comes mainly from the inter-layer interaction and multiple layers combination. As the HBP-3 already presents the best performance, thus we utilize $relu5\_1,~relu5\_2$ and $relu5\_3$ in all the experiments in Sect.~\ref{sec:comp}.
\setlength{\tabcolsep}{5pt}
\begin{table}
\begin{minipage}{\textwidth}
\centering
\caption{Quantitative analysis results on CUB-200-2011 dataset}
\begin{tabular}{c |c |c c c|c c c}
\hline
\multirow{2}{*}{Method} & \multicolumn{1}{c|}{FBP} & \multicolumn{3}{c|}{CBP} & \multicolumn{3}{c}{HBP} \\
\cline{2-8}
 & FBP-1\footnote{$relu5\_3\ast relu5\_3$.}
 & CBP-1\footnote{$relu5\_3\ast relu5\_2$.}
 & CBP-2\footnote{$relu5\_3\ast relu5\_1$.}
 & CBP-3\footnote{$relu5\_3\ast relu4\_3$.}
 & HBP-1\footnote{$relu5\_3\ast relu5\_2+relu5\_3\ast relu5\_1$.}
 & HBP-2\footnote{$relu5\_3\ast relu5\_2+relu5\_3\ast relu5\_1+relu5\_3\ast relu4\_3$.}
 & HBP-3\footnote{$relu5\_3\ast relu5\_2+relu5\_3\ast relu5\_1+relu5\_2\ast relu5\_1$.}\\ \hline

Accuracy & 85.70 & 86.75 & 86.85 & 86.67 & 86.78 & 86.91 & 87.15 \\ \hline
\end{tabular}
\label{table:table11}
\end{minipage}
\end{table}

We also compare our cross-layer integration with hypercolumn~\cite{cai2017higher} based feature fusion. For fair comparison, we re-implement hypercolumn as the feature concatenation of $relu5\_3$ and $relu5\_2$, followed by factorized bilinear pooling (denoted as HyperBP) under the same experimental settings. Table~\ref{table:table33333} shows that our CBP obtains slightly better result than HyperBP with nearly 1/2 parameters, which again indicates that our integration framework is more effective in capturing inter-layer feature relations. This is not surprising since our CBP is consistent with human perception to some extent. On the contrary of the HyperBP, which obtains even worse result when integrating more convolution layer activations~\cite{cai2017higher}, our HBP is able to capture the complementary information within intermediate convolution layers and achieves an obvious improvement in recognition accuracy.
\begin{table}
\begin{center}
\caption{Classification accuracy on the CUB dataset and model sizes of different feature integrations}
\label{table:table33333}
\begin{tabular}{|c|c|c|}
\hline
Method & Accuracy & Model Size \\
\hline\hline
HyperBP & 86.60 & 18.4M \\
\hline
CBP  & 86.75 & 10.0M \\
HBP  & $\mathbf{87.15}$ & 17.5M \\
\hline
\end{tabular}
\end{center}
\end{table}

\subsection{Comparison with State-of-the-art} \label{sec:comp}
\subsubsection{Results on CUB-200-2011.}
CUB dataset provides ground-truth annotations of bounding boxes and parts of birds. The only supervised information we use is the image level class label. The classification accuracy on CUB-200-2011 is summarized in Table~\ref{table:table2}. The table is split into three parts over the rows: the first summarizes the annotation-based methods (using object bounding boxes or part annotations); the second includes the unsupervised part-based methods; the last illustrates the results of pooling-based methods.

\begin{table}
\begin{center}
\caption{Comparison results on CUB-200-2011 dataset. Anno. represents using bounding box or part annotation}
\label{table:table2}
\begin{tabular}{|c|c|c|}
\hline
Method & Anno. & Accuracy \\
\hline\hline
SPDA-CNN~\cite{zhang2016spda} & $\surd$ & 85.1 \\
B-CNN~\cite{lin2015bilinear} & $\surd$ & 85.1 \\
PN-CNN~\cite{branson2014bird} & $\surd$ & 85.4 \\
\hline
STN~\cite{jaderberg2015spatial}  &   & 84.1 \\
RA-CNN~\cite{fu2017look} &   & 85.3 \\
MA-CNN~\cite{zheng2017learning} &   & 86.5 \\
\hline
B-CNN~\cite{lin2015bilinear}  &   & 84.0 \\
CBP~\cite{gao2016compact} &   & 84.0 \\
LRBP~\cite{kong2017low} &   & 84.2 \\
HIHCA~\cite{cai2017higher} &   & 85.3 \\
Improved B-CNN~\cite{lin2017improved}  &   & 85.8 \\
BoostCNN~\cite{moghimi2016boosted} &   & 86.2 \\
KP~\cite{cui2017kernel} &   & 86.2 \\
\hline
FBP($relu5\_3$)  &   & 85.7 \\
CBP($relu5\_3 + relu5\_2$)  &   & 86.7 \\
HBP($relu5\_3 + relu5\_2 + relu5\_1$) &   & $\mathbf{87.1}$ \\
\hline
\end{tabular}
\end{center}
\end{table}

From results in Table~\ref{table:table2}, we can see that PN-CNN~\cite{branson2014bird} uses strong supervision of both human-defined bounding box and ground truth parts. SPDA-CNN~\cite{zhang2016spda} uses ground truth parts and B-CNN~\cite{lin2015bilinear} uses bounding box with very high-dimensional feature representation (250K dimensions). The proposed HBP($relu5\_3+relu5\_2+relu5\_1$) achieves better result compared with PN-CNN~\cite{branson2014bird}, SPDA-CNN~\cite{zhang2016spda} and B-CNN~\cite{lin2015bilinear} even without bbox and part annotation, which demonstrates the effectiveness of our model. Compared with STN~\cite{jaderberg2015spatial} which uses stronger inception network as baseline model, we obtain a relative accuracy gain with 3.6\% by our HBP($relu5\_3+relu5\_2+relu5\_1$). We even surpass RA-CNN~\cite{fu2017look} and MA-CNN~\cite{zheng2017learning}, which are the recently-proposed state-of-the-art unsupervised part-based methods, with 2.1\% and 0.7\% relative accuracy gains, respectively. Compared with the baselines of pooling-based model B-CNN~\cite{lin2015bilinear}, CBP~\cite{gao2016compact} and LRBP~\cite{kong2017low}, the superior result that we achieve mainly benefits from the inter-layer interaction of feature and the integration of multiple layers. We also surpass BoostCNN~\cite{moghimi2016boosted} which boosts multiple bilinear networks trained at multiple scales. Although HIHCA~\cite{cai2017higher} proposes similar ideas to model feature interaction for fine-grained recognition, our model can achieve higher accuracy because of the mutual reinforcement framework for inter-layer feature interaction and discriminative feature learning. Note that HBP($relu5\_3+relu5\_2+relu5\_1$) outperforms CBP($relu5\_3+relu5\_2$) and FBP($relu5\_3$), which indicates that our model can capture the complementary information among layers.

\subsubsection{Results on Stanford Cars.}
The classification accuracy on Stanford Cars is summarized in Table~\ref{table:table3}. Different car parts are discriminative and complementary, thus object and part localization may play a significant role here~\cite{yang2015large}. Although our HBP has no explicit part detection, we achieve the best result among state-of-the-art methods. Relying on inter-layer feature interaction learning, we even surpass PA-CNN~\cite{krause2015fine} by 1.2\% relative accuracy gains, which uses human-defined bounding box. We can observe significant improvement compared with unsupervised part-based method MA-CNN~\cite{zheng2017learning}. Our HBP is also better than pooling-based methods BoostCNN~\cite{moghimi2016boosted} and KP~\cite{cui2017kernel}.
\begin{table}
\begin{center}
\caption{Comparison results on Stanford Cars dataset. Anno. represents using bounding box}
\label{table:table3}
\begin{tabular}{|c|c|c|}
\hline
Method & Anno. & Accuracy \\
\hline\hline
FCAN~\cite{liu2016fully} & $\surd$ & 91.3 \\
PA-CNN~\cite{krause2015fine} & $\surd$ & 92.6 \\
\hline
FCAN~\cite{liu2016fully}  &   & 89.1 \\
RA-CNN~\cite{fu2017look} &   & 92.5 \\
MA-CNN~\cite{zheng2017learning} &   & 92.8 \\
\hline
B-CNN~\cite{lin2015bilinear}  &   & 90.6 \\
LRBP~\cite{kong2017low} &   & 90.9 \\
HIHCA~\cite{cai2017higher} &   & 91.7 \\
Improved B-CNN~\cite{lin2017improved}  &   & 92.0 \\
BoostCNN~\cite{moghimi2016boosted} &   & 92.1 \\
KP~\cite{cui2017kernel} &   & 92.4 \\
\hline
HBP &   & $\mathbf{93.7}$ \\
\hline
\end{tabular}
\end{center}
\end{table}

\subsubsection{Results on FGVC-Aircraft.}
Different aircraft models are difficult to be recognized, due to subtle differences, e.g., one may be able to distinguish them by counting the number of windows in the model. The classification accuracy on FGVC-Aircraft is summarized in Table~\ref{table:table4}. Still, our model achieves the highest classification accuracy among all the methods. We can observe stable improvement compared with annotation-based method MDTP~\cite{wang2016mining}, part learning-based method MA-CNN~\cite{zheng2017learning}, and pooling-based BoostCNN~\cite{moghimi2016boosted}, which highlights the efficacy and robustness of the proposed HBP model.
\begin{table}
\begin{center}
\caption{Comparison results on FGVC-Aircraft dataset. Anno. represents using bounding box}
\label{table:table4}
\begin{tabular}{|c|c|c|}
\hline
Method & Anno. & Accuracy \\
\hline\hline
MG-CNN~\cite{wang2015multiple}  & $\surd$ & 86.6 \\
MDTP~\cite{wang2016mining} & $\surd$ & 88.4 \\
\hline
RA-CNN~\cite{fu2017look} &   & 88.2 \\
MA-CNN~\cite{zheng2017learning} &   & 89.9 \\
\hline
B-CNN~\cite{lin2015bilinear}  &   & 86.9 \\
KP~\cite{cui2017kernel} &   & 86.9 \\
LRBP~\cite{kong2017low} &   & 87.3 \\
HIHCA~\cite{cai2017higher} &   & 88.3 \\
Improved B-CNN~\cite{lin2017improved}  &   & 88.5 \\
BoostCNN~\cite{moghimi2016boosted} &   & 88.5 \\
\hline
HBP &   & $\mathbf{90.3}$ \\
\hline
\end{tabular}
\end{center}
\end{table}

\subsection{Qualitative Visualization} \label{sec:qual}

\setlength{\tabcolsep}{0.5pt}
\begin{figure} 
    \centering
    \begin{tabular}{ccccccc}

        \includegraphics[width=0.14\textwidth]{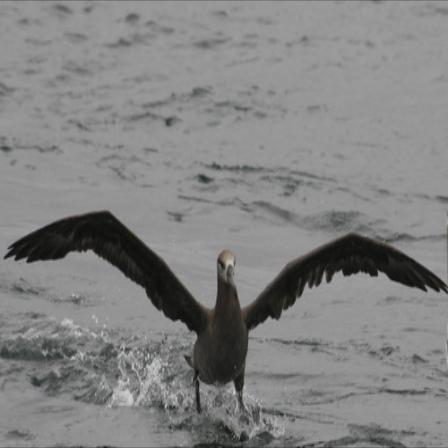} & \includegraphics[width=0.14\textwidth]{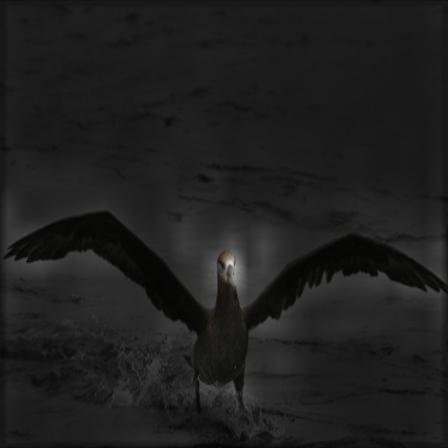} & \includegraphics[width=0.14\textwidth]{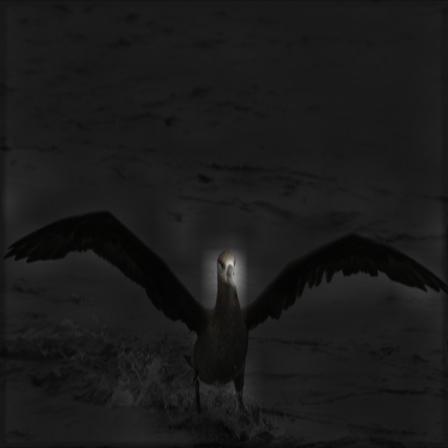} & \includegraphics[width=0.14\textwidth]{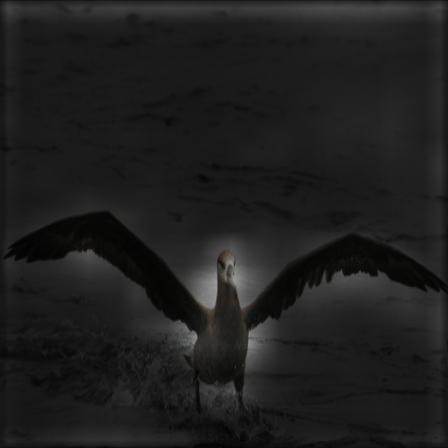} &\includegraphics[width=0.14\textwidth]{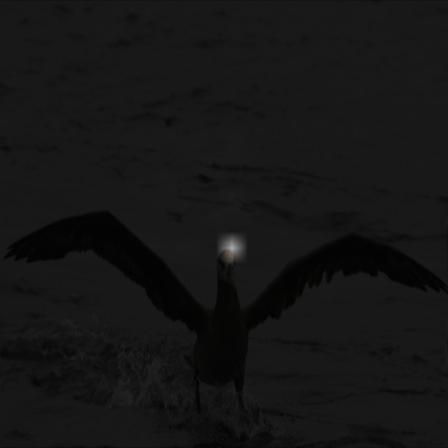} & \includegraphics[width=0.14\textwidth]{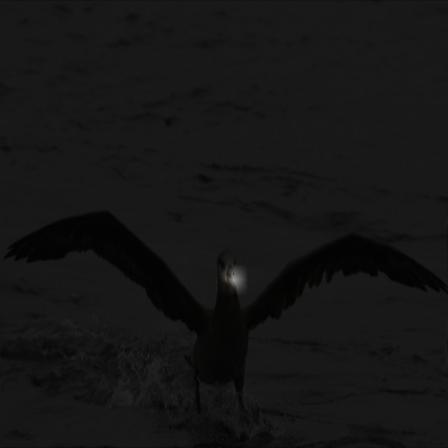} &\includegraphics[width=0.14\textwidth]{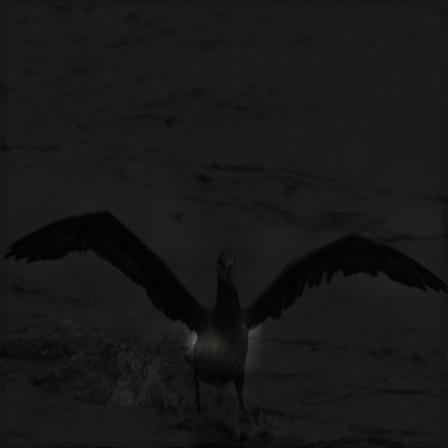} \\ [-2pt]
        \includegraphics[width=0.14\textwidth]{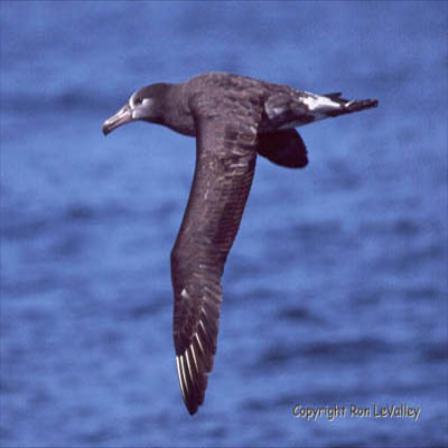} & \includegraphics[width=0.14\textwidth]{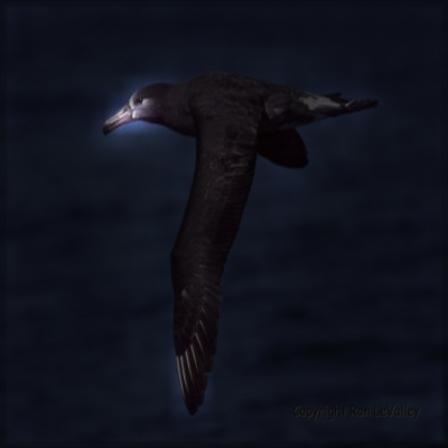} & \includegraphics[width=0.14\textwidth]{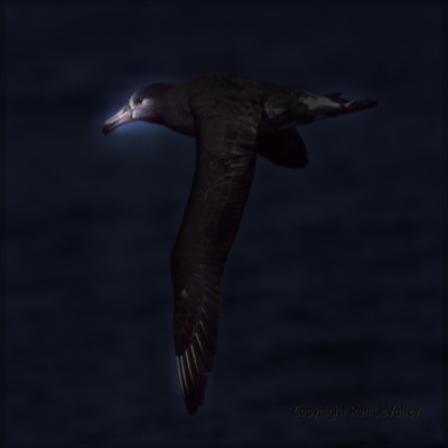} & \includegraphics[width=0.14\textwidth]{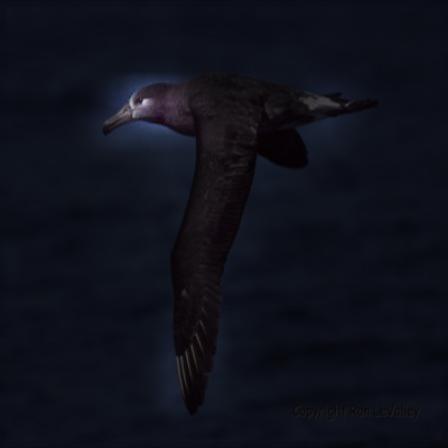} &\includegraphics[width=0.14\textwidth]{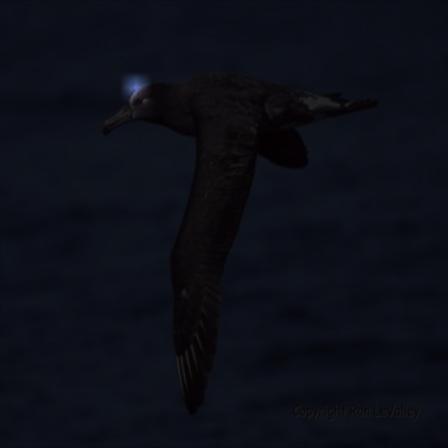} & \includegraphics[width=0.14\textwidth]{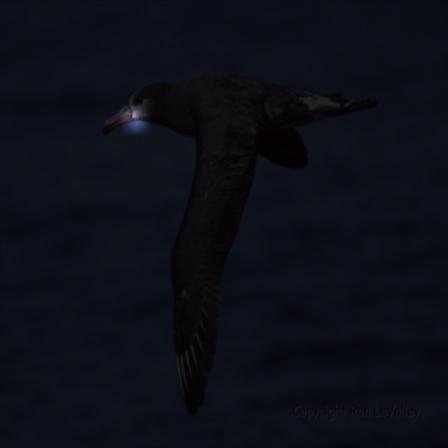} &\includegraphics[width=0.14\textwidth]{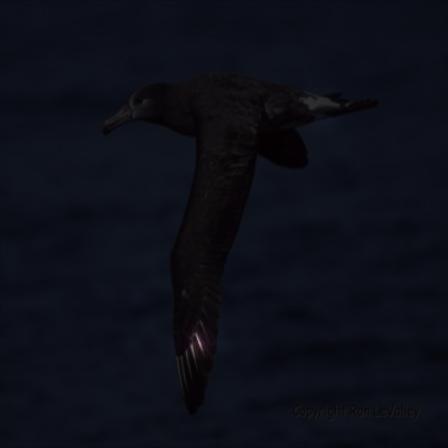} \\ [-2pt]
        \includegraphics[width=0.14\textwidth]{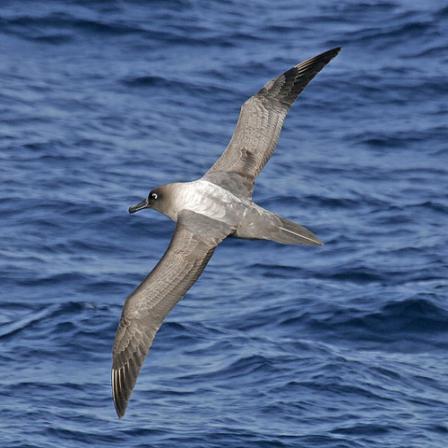} & \includegraphics[width=0.14\textwidth]{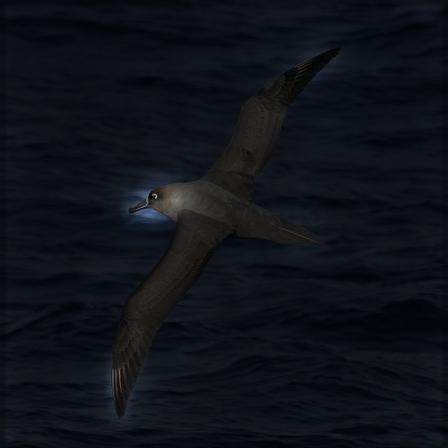} & \includegraphics[width=0.14\textwidth]{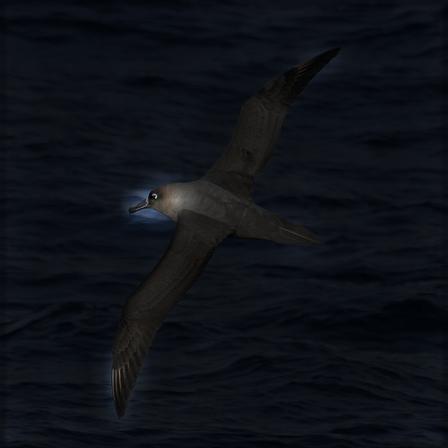} & \includegraphics[width=0.14\textwidth]{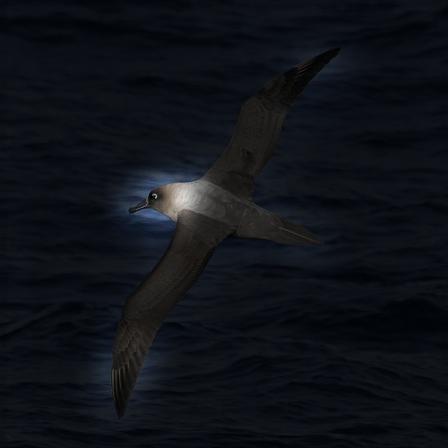} &\includegraphics[width=0.14\textwidth]{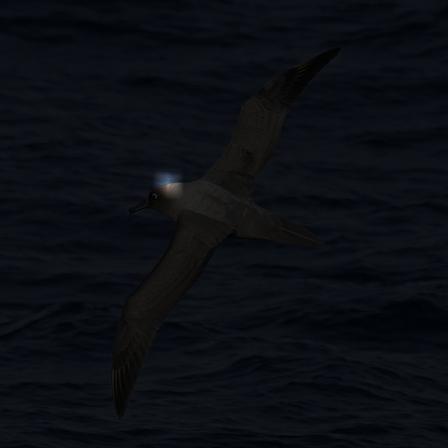} & \includegraphics[width=0.14\textwidth]{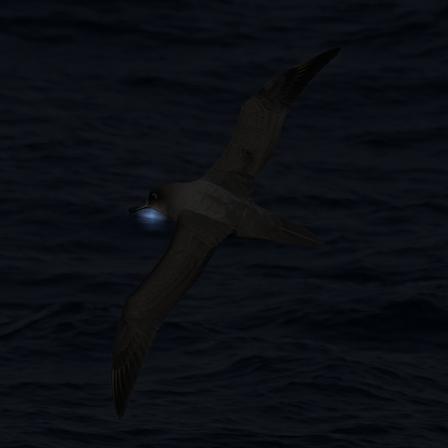} &\includegraphics[width=0.14\textwidth]{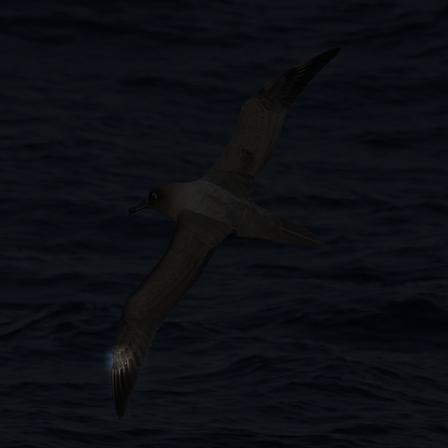} \\ [5pt]

        \includegraphics[width=0.14\textwidth]{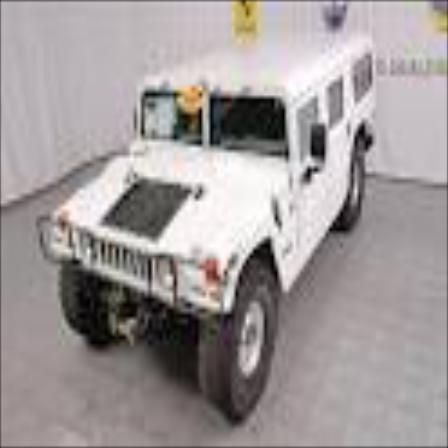} & \includegraphics[width=0.14\textwidth]{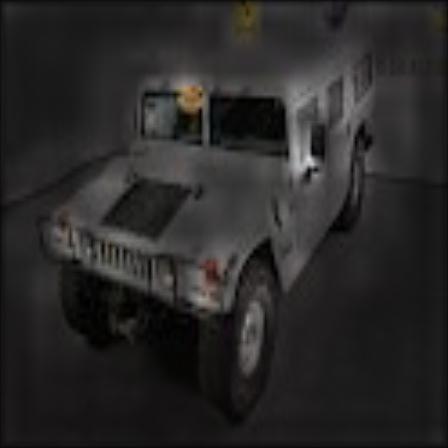} & \includegraphics[width=0.14\textwidth]{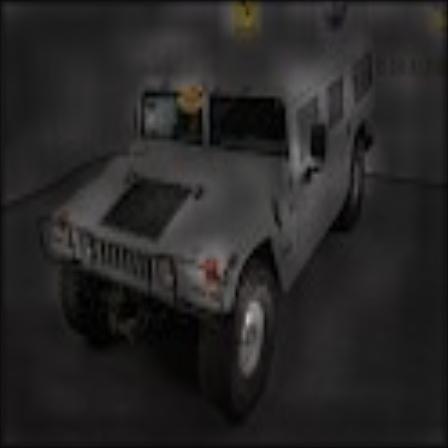} & \includegraphics[width=0.14\textwidth]{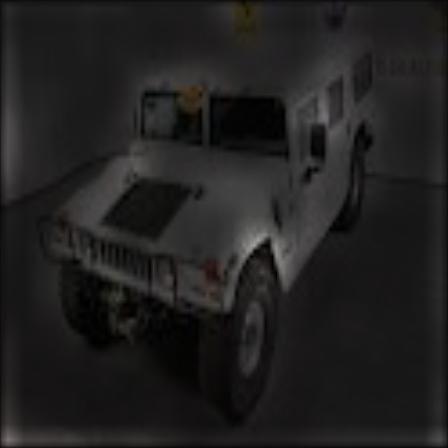} &\includegraphics[width=0.14\textwidth]{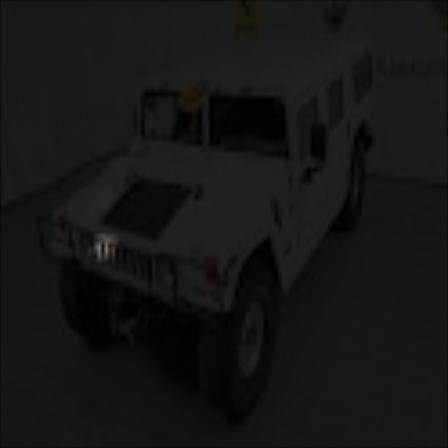} & \includegraphics[width=0.14\textwidth]{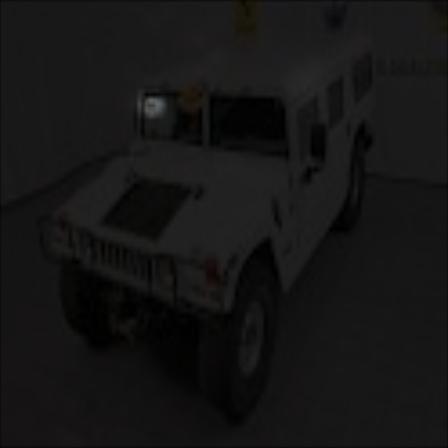} &\includegraphics[width=0.14\textwidth]{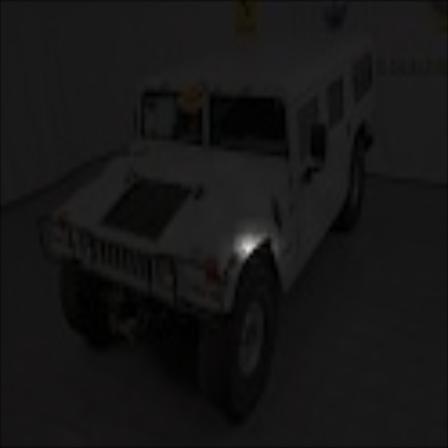} \\ [-2pt]
        \includegraphics[width=0.14\textwidth]{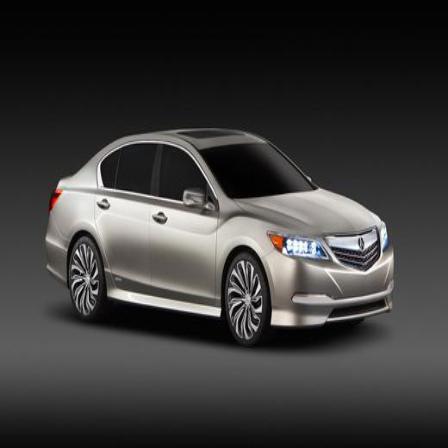} & \includegraphics[width=0.14\textwidth]{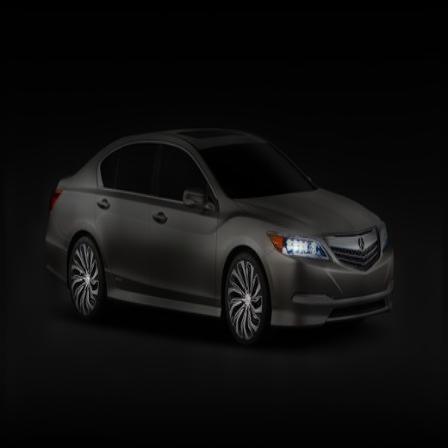} & \includegraphics[width=0.14\textwidth]{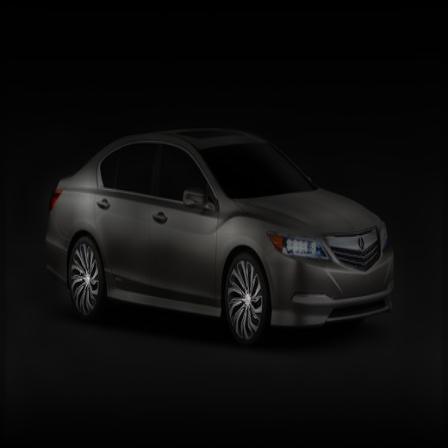} & \includegraphics[width=0.14\textwidth]{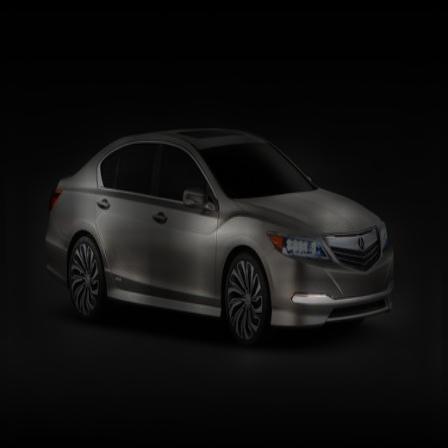} &\includegraphics[width=0.14\textwidth]{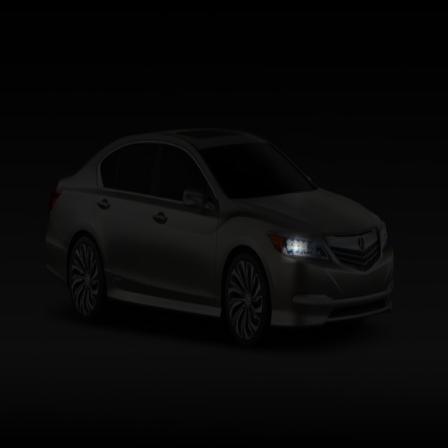} & \includegraphics[width=0.14\textwidth]{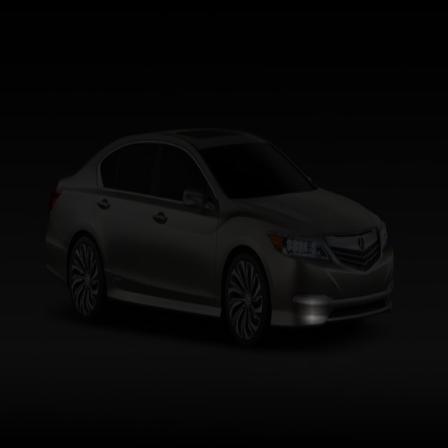} &\includegraphics[width=0.14\textwidth]{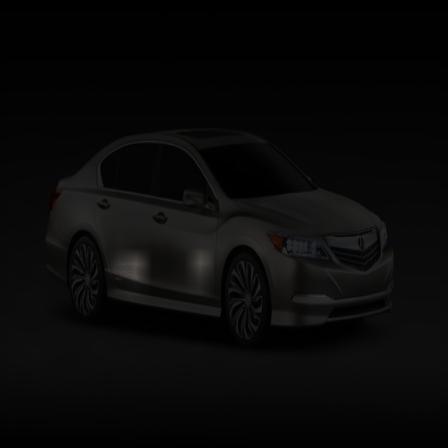} \\ [-2pt]
        \includegraphics[width=0.14\textwidth]{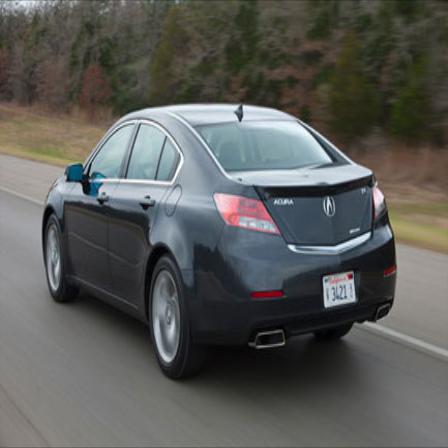} & \includegraphics[width=0.14\textwidth]{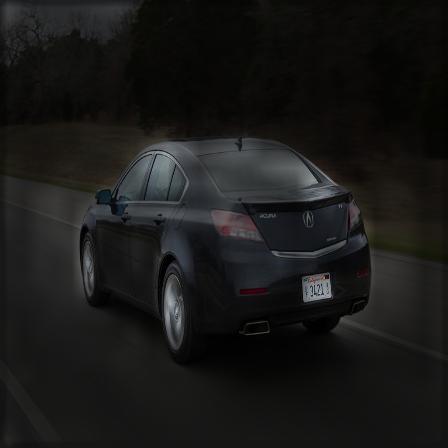} & \includegraphics[width=0.14\textwidth]{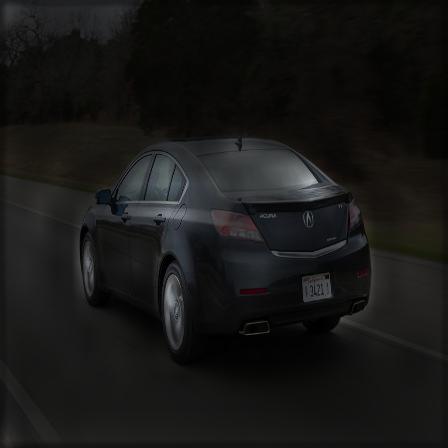} & \includegraphics[width=0.14\textwidth]{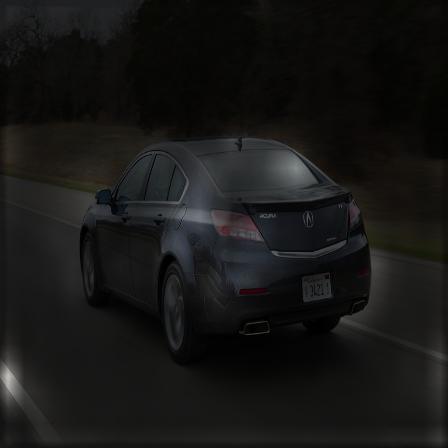} &\includegraphics[width=0.14\textwidth]{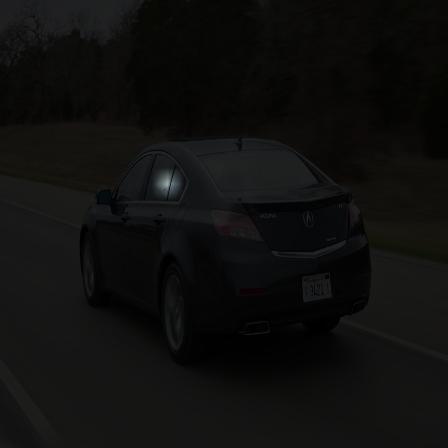} & \includegraphics[width=0.14\textwidth]{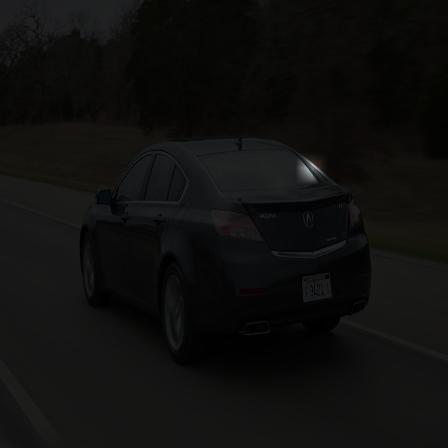} &\includegraphics[width=0.14\textwidth]{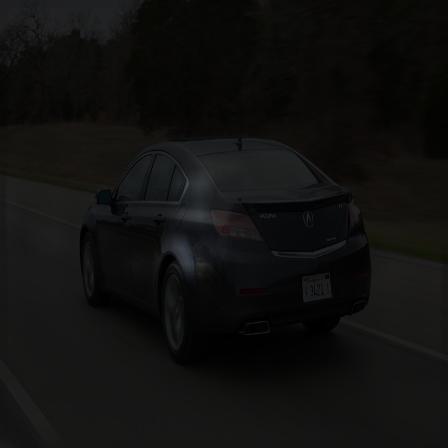} \\ [5pt]

        \includegraphics[width=0.14\textwidth]{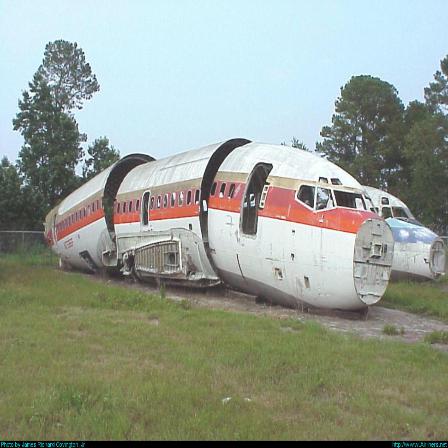} & \includegraphics[width=0.14\textwidth]{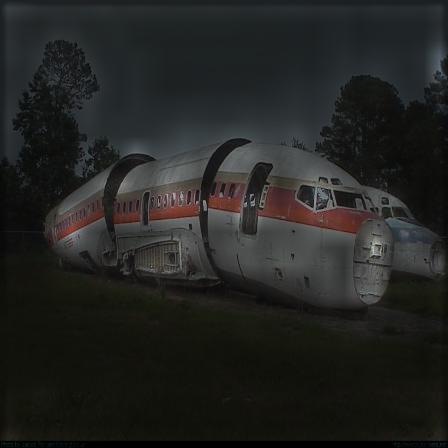} & \includegraphics[width=0.14\textwidth]{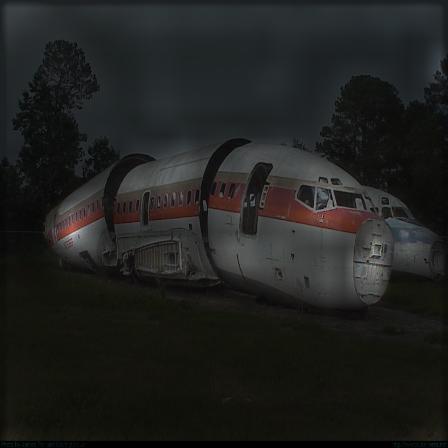} & \includegraphics[width=0.14\textwidth]{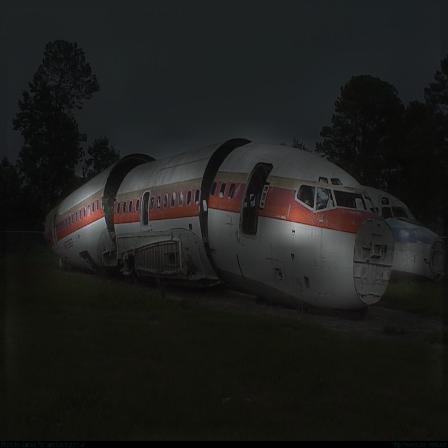} &\includegraphics[width=0.14\textwidth]{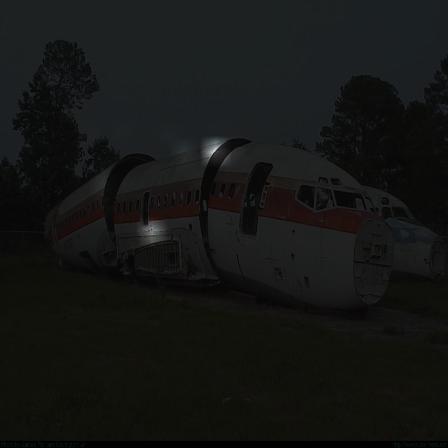} & \includegraphics[width=0.14\textwidth]{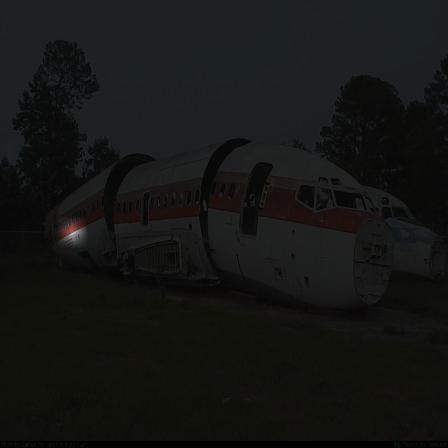} &\includegraphics[width=0.14\textwidth]{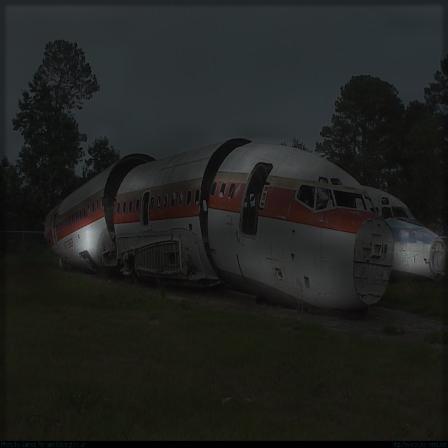} \\ [-2pt]
        \includegraphics[width=0.14\textwidth]{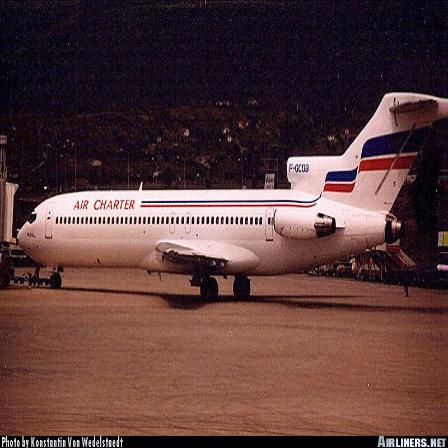} & \includegraphics[width=0.14\textwidth]{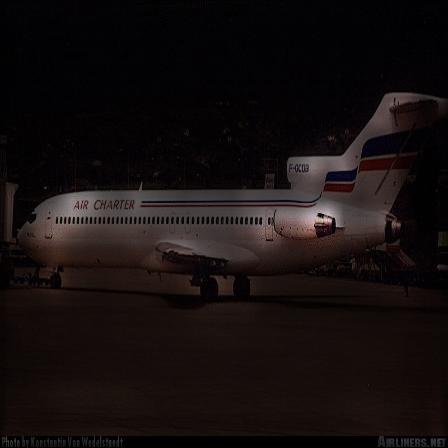} & \includegraphics[width=0.14\textwidth]{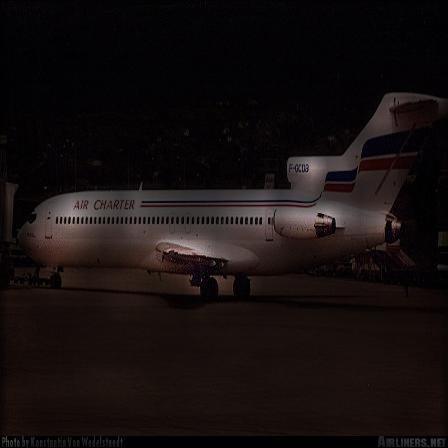} & \includegraphics[width=0.14\textwidth]{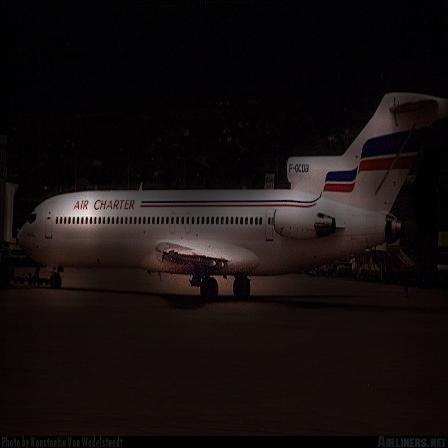} &\includegraphics[width=0.14\textwidth]{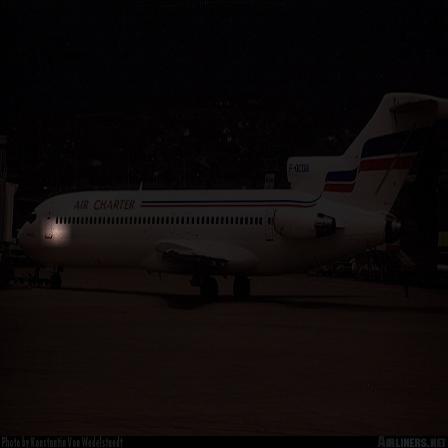} & \includegraphics[width=0.14\textwidth]{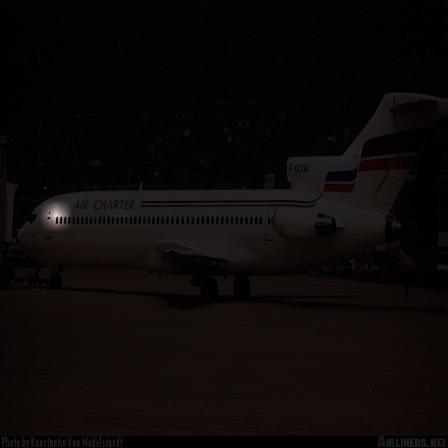} &\includegraphics[width=0.14\textwidth]{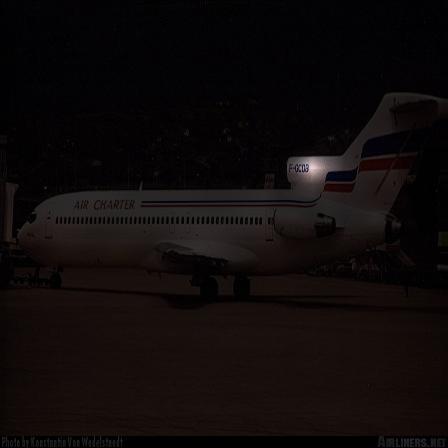} \\ [-2pt]
        \includegraphics[width=0.14\textwidth]{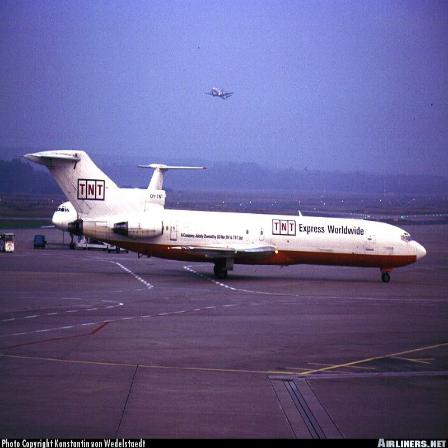} & \includegraphics[width=0.14\textwidth]{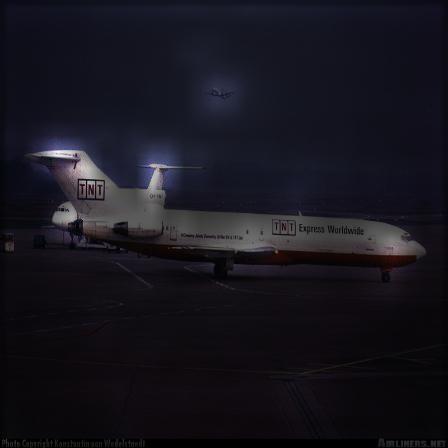} & \includegraphics[width=0.14\textwidth]{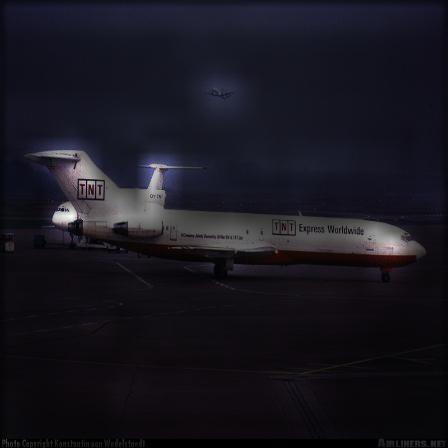} & \includegraphics[width=0.14\textwidth]{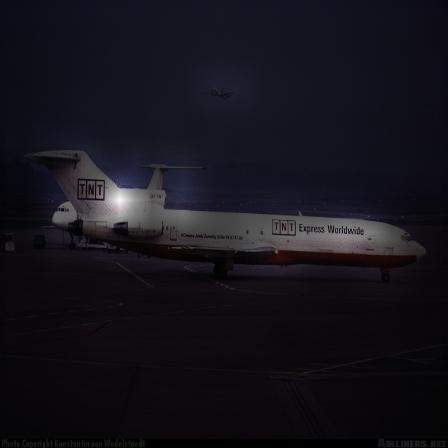} &\includegraphics[width=0.14\textwidth]{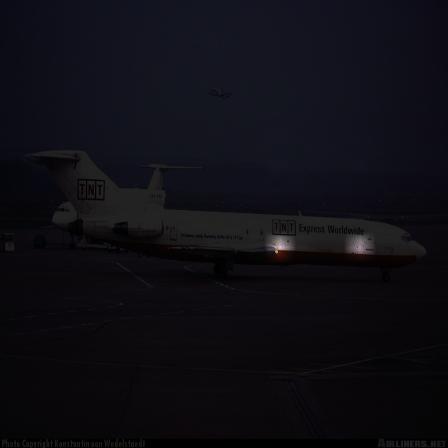} & \includegraphics[width=0.14\textwidth]{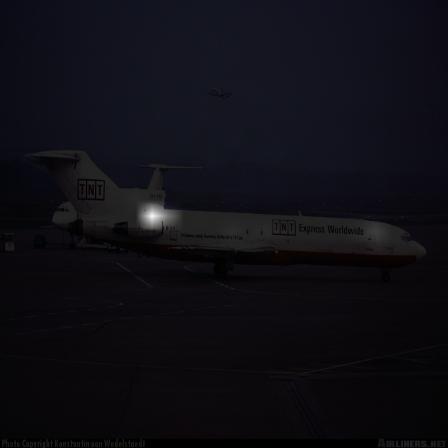} &\includegraphics[width=0.14\textwidth]{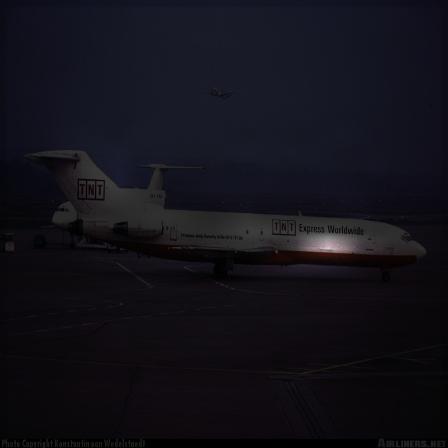} \\ [5pt]

         $Original$ & $relu5\_1$ & $relu5\_2$ & $relu5\_3$ & $project5\_1$ & $project5\_2$ & $project5\_3$
    \end{tabular}
    \caption{Visualization of model response of different layers on the CUB, Cars and Aircraft datasets. It can be seen that our model tend to ignore features in the cluttered background and focus on the most discriminative parts of object.}
    \label{fig:fig3}
\end{figure}

To better understand our model, we visualize the model response of different layers in our fine-tuned network on different datasets. We obtain the activation maps by computing the magnitude of feature activations averaged across channel. In Fig.~\ref{fig:fig3}, we show some randomly selected images from three different datasets and their corresponding visualizations.

The visualizations all suggest that the proposed model is capable of ignoring cluttered backgrounds and tends to activate strongly on highly specific semantic parts. The highlighted activation regions in $project5\_1,~project5\_2$ and $project5\_3$ are strongly related to semantic parts, such as heads, wings and breast in CUB; front bumpers, wheels and lights in Cars; cockpit, tail stabilizers and engine in Aircraft. These parts are crucial to distinguish the category. Moreover, our model is highly consistent with the human perception that resolve the fine details when perceive scenes or objects. In Fig.~\ref{fig:fig3}, we can see that the convolution layers ($relu5\_1,relu5\_2,relu5\_3$) provide a rough localization of target object. Based on this, the projection layers ($project5\_1,project5\_2,project5\_3$) further determine essential parts of the object, which distinguish its category by successive interaction and integration of different part features. The process is consistent with the coarse-to-fine nature of human perception~\cite{lu2018revealing} inspired by the Gestalt dictum that the ``whole'' is prior to the ``parts'' and it also provides an intuitive explanation as to why our framework can model subtle and local differences between subcategories without explicit part detection.

\section{Conclusions}
\label{sec:conc}
In this paper, we propose a hierarchical bilinear pooling approach to fuse multi-layer features for fine-grained recognition, which combines inter-layer interactions and discriminative feature learning in a mutually-reinforced way. The proposed network requires no bounding box/part annotations and can be trained end-to-end. Extensive experiments on birds, cars and aircrafts demonstrate the effectiveness of our framework. In the future, we will conduct extended research on two directions, i.e., how to effectively fuse more layer features to obtain part representation at multiple scales, and how to merge effective methods for parts localization to learn better fine-grained representation.

\subsubsection*{Acknowledgements.}
This work was supported in part by the National Natural Science Foundation of China (No.61772220, 61571205), in part by National Key Technology Research and Development Program of Ministry of Science and Technology of China (No.2015BAK36B00), in part by the Technology Innovation Program of Hubei Province (No.2017AAA017), in part by the Key Program for International S\&T Cooperation Projects of China (No.2016YFE0121200).
%
%
%
\bibliographystyle{splncs04}
\bibliography{samplepaper}
%
%
%
%
%
\end{document}